\documentclass[11pt,a4paper]{article}
\usepackage{emnlp2017}
\usepackage{times}

\usepackage{url}
\usepackage{latexsym}
\usepackage{adjustbox,lipsum}
\usepackage{amsmath}

\usepackage{xspace}
\usepackage{xargs} 
\usepackage[colorinlistoftodos,prependcaption,textsize=tiny]{todonotes}
\usepackage{marginnote}
\newcommandx{\todoir}[2][1=]{\todo[inline]{SR: #2}\xspace}
\newcommandx{\todosr}[2][1=]{\todo[linecolor=red,backgroundcolor=red!25,bordercolor=red,#1]{SR: #2}\xspace}
\newcommandx{\todomn}[2][1=]{\todo[linecolor=cyan,backgroundcolor=cyan!25,bordercolor=cyan,#1]{MN: #2}\xspace}
\newcommandx{\todotm}[2][1=]{\todo[linecolor=blue,backgroundcolor=blue!10,bordercolor=blue,#1]{TM: #2}\xspace}

\usepackage{hyperref}
\hypersetup{
    colorlinks,
    linkcolor={blue!25!black},
    citecolor={blue!25!black},
    urlcolor={blue!25!black}
}

\usepackage{subcaption}
\usepackage{cleveref}
\Crefformat{equation}{(#2#1#3)}

\usepackage{ccg}

\emnlpfinalcopy

\def\emnlppaperid{***}

\newcommand\bleu{\textsc{bleu}\xspace}

\usepackage{url}
\usepackage{array}
\usepackage{verbatim}
\usepackage{tikz}
\usetikzlibrary{shapes}
\usepackage{tikz-qtree}
\usepackage{fixltx2e}
\usepackage{comment}
\usepackage{multirow}
\usepackage{breqn}
\usepackage{enumitem}

\usepackage{tikz-dependency}

\title{Predicting Target Language CCG Supertags Improves Neural Machine Translation}

\author{ Maria N\u{a}dejde$^{1}$ \and Siva Reddy$^{1}$  \and Rico Sennrich$^{1}$ \and Tomasz Dwojak$^{1,2}$  \\ 
 {\bf Marcin Junczys-Dowmunt$^{2}$}  \and {\bf Philipp Koehn}$^{3}$ \and {\bf Alexandra Birch}$^{1}$\\
$^{1}$School of Informatics, University of Edinburgh \\
$^{2}$Adam Mickiewicz University\\ 
 $^{3}$Dep. of Computer Science, Johns Hopkins University \\
 {\tt \{m.nadejde,siva.reddy, rico.sennrich, a.birch\}@ed.ac.uk} \\
   {\tt \{t.dwojak,junczys\}@amu.edu.pl}, {\tt phi@jhu.edu } \\
   }

\date{}

\begin{document}
\maketitle

\begin{abstract}
  
Neural machine translation (NMT) models are able to partially learn syntactic information from sequential lexical information. Still, some complex syntactic phenomena such as prepositional phrase attachment are poorly modeled. 
This work aims to answer two questions: 1) Does explicitly modeling target language syntax help NMT? 2) Is tight integration of words and syntax better than multitask training?
We introduce syntactic information in the form of CCG supertags in the decoder, by interleaving the target supertags with the word sequence.
Our results on WMT data show that explicitly modeling target-syntax improves machine translation quality for German$\rightarrow$English, a high-resource pair, and for Romanian$\rightarrow$English, a low-resource pair and also several syntactic phenomena including prepositional phrase attachment. Furthermore, a tight coupling of words and syntax improves translation quality more than multitask training. By combining target-syntax with adding source-side dependency labels in the embedding layer, we obtain a total improvement of  0.9 \bleu for German$\rightarrow$English and 1.2 \bleu for Romanian$\rightarrow$English.
\end{abstract}

\section{Introduction}
\label{intro}

Sequence-to-sequence neural machine translation (NMT) models~\citep{Sutskever2014,cho-al-emnlp14,bahdanau2015neural} are state-of-the-art on a multitude of language-pairs \citep{sennrich-haddow-birch:2016:WMT, junczys2016neural}. Part of the appeal of neural models is that they can learn to implicitly model phenomena which underlie high quality output, and some syntax is indeed captured by these models. In a detailed analysis,  \citet{BentivogliBCF16} show that NMT significantly improves over phrase-based SMT, in particular with respect to morphology and word order, but that results can still be improved for longer sentences and complex syntactic phenomena such as prepositional phrase (PP) attachment.
 Another study by \citet{shi-padhi-knight:2016:EMNLP2016} shows that the encoder layer of NMT partially learns syntactic information about the source language, however complex syntactic phenomena such as coordination or PP attachment are poorly modeled. 

Recent work which incorporates additional source-side linguistic information in NMT models~\citep{luong2015multi,sennrich2016linguistic} show that
even though neural models have strong learning capabilities, 
explicit features can still improve translation quality.
In this work, we examine the benefit of incorporating global syntactic information on the target-side. We also address the question of how best to incorporate this information.
For language pairs where syntactic resources are available on both the source and target-side, we show that approaches to incorporate source syntax and target syntax are complementary. 

We propose a method for tightly coupling words and syntax by interleaving the target syntactic representation with the word sequence. We compare this to loosely coupling words and syntax using a multitask solution, where the shared parts of the model are trained to produce either a target sequence of words or supertags in a similar fashion to ~\citet{luong2015multi}. 

We use CCG syntactic categories \citep{steedman2000syntactic}, also known as \textit{supertags}, to represent syntax explicitly. Supertags provide global syntactic information locally at the lexical level.
They encode subcategorization information, capturing short and long range dependencies and attachments, and also tense and morphological aspects of the word in a given context. 
Consider the sentence in Figure~\ref{figure:SourceCCGRepresentation}. 
This sentence contains two PP attachments and could lead to several disambiguation possibilities (\textsl{``in''} can attach to \textsl{``Netanyahu"} or \textsl{``receives"}, and \textsl{``of"} can attach to \textsl{``capital"}, \textsl{``Netanyahu"} or \textsl{``receives"}). These alternatives may lead to different translations in other languages. However the supertag \cf{((S[dcl]\backslash NP)/PP)/NP} of  \textsl{``receives"} indicates that the preposition \textsl{``in"} attaches to the verb, and the supertag \cf{(NP\backslash NP) /NP} of \textsl{``of"} indicates that it attaches to \textsl{``capital"}, thereby resolving the ambiguity.

\begin{figure*}[th]
\begin{center}
\begin{adjustbox}{max width=1\textwidth}
\renewcommand{\arraystretch}{1.2}
\begin{tabular}{ l  c  c  c  c  c  c  c  c  c  c }

\multicolumn{11}{l}{ \textbf{Source-side } } \\ \hline
BPE: & Obama & receives & Net+ & an+ & yahu & in &  the  & capital & of & USA \\
IOB: & O & O & B &  I & E & O & O & O & O & O \\
CCG: & \cf{NP} & \cf{((S[dcl]\backslash NP)/PP)/NP} & \cf{NP} & \cf{NP} & \cf{NP} & \cf{PP/NP} &  \cf{NP/N}  & \cf{N} & \cf{(NP\backslash NP)/NP} & \cf{NP}   \\

\multicolumn{11}{l}{ \textbf{Target-side } } \\ \hline
\multicolumn{11}{c}{  \cf{NP} \ Obama \  \cf{((S[dcl]\backslash NP)/PP)/NP} \ receives \ \cf{NP} \  Net+ \  an+ \  yahu \ \cf{PP/NP} \ in  \ \cf{NP/N}\  the \ \cf{N} \  capital \  \cf{(NP\backslash NP)/NP} \  of \  \cf{NP} \ USA} \\
\end{tabular}
\end{adjustbox}
\end{center}
\caption{\label{figure:SourceCCGRepresentation} Source and target representation of syntactic information in syntax-aware NMT. } 
\end{figure*}

Our research contributions are as follows:
\vspace{-0.3em}
\begin{itemize}[leftmargin=*]
\item We propose a novel approach to integrating target syntax at word level in the decoder, by interleaving CCG supertags in the target word sequence. 
\item We show that the target language syntax improves translation quality for German$\rightarrow$English and Romanian$\rightarrow$English as measured by BLEU.
Our results suggest that a tight coupling of target words and syntax (by interleaving) improves translation quality more than the decoupled signal from multitask training.
\item We show that incorporating source-side linguistic information is complimentary to our method, further improving the translation quality.
\item We present a fine-grained analysis of SNMT and show consistent gains for different linguistic phenomena and sentence lengths.
\end{itemize}

\section{Related work}
\label{section:related}

Syntax has helped in statistical machine translation (SMT) to capture dependencies between distant words that impact morphological agreement, subcategorisation and word order \citep{GalleyEtAl2004,Quirk2007, Williams2012, NadejdeEtAl2013, sennrich15, NadejdeEtAl2016, NadejdeEtAl2016b,Chiang2007}. There has been some work in NMT on modeling source-side syntax implicitly or explicitly. \citet{kalchbrenner-blunsom:2013:EMNLP,cho-EtAl:2014:SSST-8} capture the hierarchical aspects of language implicitly by using convolutional neural networks, while \citet{eriguchi-hashimoto-tsuruoka:2016} use the parse tree of the source sentence to guide the recurrence and attention model in tree-to-sequence NMT.
 \citet{luong2015multi} co-train a translation model and a source-side syntactic parser which share the encoder. Our multitask models extend their work
to attention-based NMT models and to predicting target-side syntax as the secondary task.
\citet{sennrich2016linguistic} generalize the embedding layer of NMT to include explicit linguistic features such as dependency relations and part-of-speech tags and we use their framework to show source and target syntax provide complementary information.

Applying more tightly coupled linguistic factors on the target for NMT has been previously investigated.
\citet{niehuesusing} proposed a factored RNN-based language model for re-scoring an n-best list produced by a phrase-based MT system. In recent work, 
\citet{martinez2016factored} implemented a factored NMT decoder which generated both lemmas and morphological tags. The two factors were then post-processed to generate the word form. Unfortunately no real gain was reported for these experiments. Concurrently with our work, \citet{Aharoni2017} proposed serializing the target constituency trees and \citet{Akiko2017} model target dependency relations by augmenting the NMT decoder with a RNN grammar \citep{dyer-EtAl:2016:N16-1}. In our work, we use CCG supertags which are a more compact representation of global syntax.  Furthermore, we do not focus on model architectures, and instead we explore the more general problem of including target syntax in NMT: comparing tightly and loosely coupled syntactic information and showing source and target syntax are complementary.

Previous work on integrating CCG supertags in factored phrase-based models \citep{BirchCCG} made strong independence assumptions between the target word sequence and the CCG categories. 
In this work we take advantage of the expressive power of recurrent neural networks to learn representations that generate both words and CCG supertags, conditioned on the entire lexical and syntactic target history.

\section{Modeling Syntax in NMT}
\label{section:SyntacticRep}

CCG is a lexicalised formalism in which words are assigned with syntactic categories, i.e., supertags, that indicate context-sensitive morpho-syntactic properties of a word in a sentence. The combinators of CCG allow the supertags to capture global syntactic constraints locally. Though NMT captures long range dependencies using long-term memory, short-term memory is cheap and reliable. Supertags can help by allowing the model to rely more on local information (short-term) and not having to rely heavily on long-term memory.

Consider a decoder that has to generate the following sentences:

\begin{enumerate}[leftmargin=*]
\item What$_{(S[wq]/(S[q]/NP))/N}$ city is$_{(S[q]/PP)/NP}$ the Taj Mahal in? 
\item Where$_{S[wq]/(S[q]/NP)}$ is$_{(S[q]/NP)/NP}$ the Taj Mahal?
\end{enumerate}

If the decoding starts with predicting ``\textsl{What}'', it is ungrammatical to omit the preposition ``\textsl{in}'', and if the decoding starts with predicting ``\textsl{Where}'', it is ungrammatical to predict the preposition. Here the decision to predict ``\textsl{in}'' depends on the first word, a long range dependency. However if we rely on CCG supertags, the supertags of both these sequences look very different. The supertag \cf{(S[q]/PP)/NP} for the verb ``\textsl{is}'' in the first sentence indicates that a preposition is expected in future context. Furthermore it is likely to see this particular supertag of the verb in the context of \cf{(S[wq]/(S[q]/NP))/N} but it is unlikely in the context of \cf{S[wq]/(S[q]/NP)}. Therefore a succession of local decisions based on CCG supertags will result in the correct prediction of the preposition in the first sentence, and omitting the preposition in the second sentence. Since the vocabulary of CCG supertags is much smaller than that of possible words, the NMT model will do a better job at generalizing over and predicting the correct CCG supertags sequence. 
 
CCG supertags also help during encoding if they are given in the input, as we saw with the case of PP attachment in \Cref{figure:SourceCCGRepresentation}. Translation of the correct verb form and agreement can be improved with CCG since supertags also encode tense, morphology and agreements. For example, in the sentence ``\textsl{It is going to rain}'', the supertag \cf{(S[ng]\backslash NP[expl])/(S[to]\backslash NP)} of ``\textsl{going}'' indicates the current word is a verb in continuous form looking for an infinitive construction on the right, and an expletive pronoun on the left.  

We explore the effect of target-side syntax by using CCG supertags in the decoder and by combining these with source-side syntax in the encoder, as follows.

\paragraph{Baseline decoder} 
The baseline decoder architecture is a conditional GRU with attention ($cGRU_{attn}$) as implemented in the Nematus toolkit \citep{sennrich-EtAl:2017}. The decoder is a recursive function computing a hidden state $s_{j}$ at each time step $j \in [1,T]$ of the target recurrence. This function takes as input the previous hidden state $s_{j-1}$, the embedding of the previous target word $y_{j-1}$ and the output of the attention model $c_{j}$. The attention model computes a weighted sum over the hidden states $h_{i} = [\overrightarrow{h_{i}};\overleftarrow{h_{i}}]$ of the bi-directional RNN encoder. The function $g$ computes the intermediate representation $t_{j}$ and passes this to a \textit{softmax} layer which first applies a linear transformation ($W_o$) and then computes the probability distribution over the target vocabulary. The training objective for the entire architecture is minimizing the discrete cross-entropy, therefore the loss $l$ is the negative log-probability of the reference sentence. 

\begin{align}
& s_{j}' = GRU_{1}(y_{j-1},s_{j-1}) \\
& c_j = ATT ([h_1; ... ;h_{|x|}], s_{j}')\\
& s_j = cGRU_{attn}(y_{j-1},s_{j-1}, c_j) \\
& t_j = g(y_{j-1},s_{j}, c_j) \\
& p_y = \prod_{j=1}^{T} p(y_{j} | x,y_{1:j-1}) = \prod_{j=1}^{T} softmax(t_j W_o) \\
& l = -  log(p_y) 
\end{align}


\begin{figure*}[ht]
\begin{tabular}{c c}

  \includegraphics[width=0.8\columnwidth]{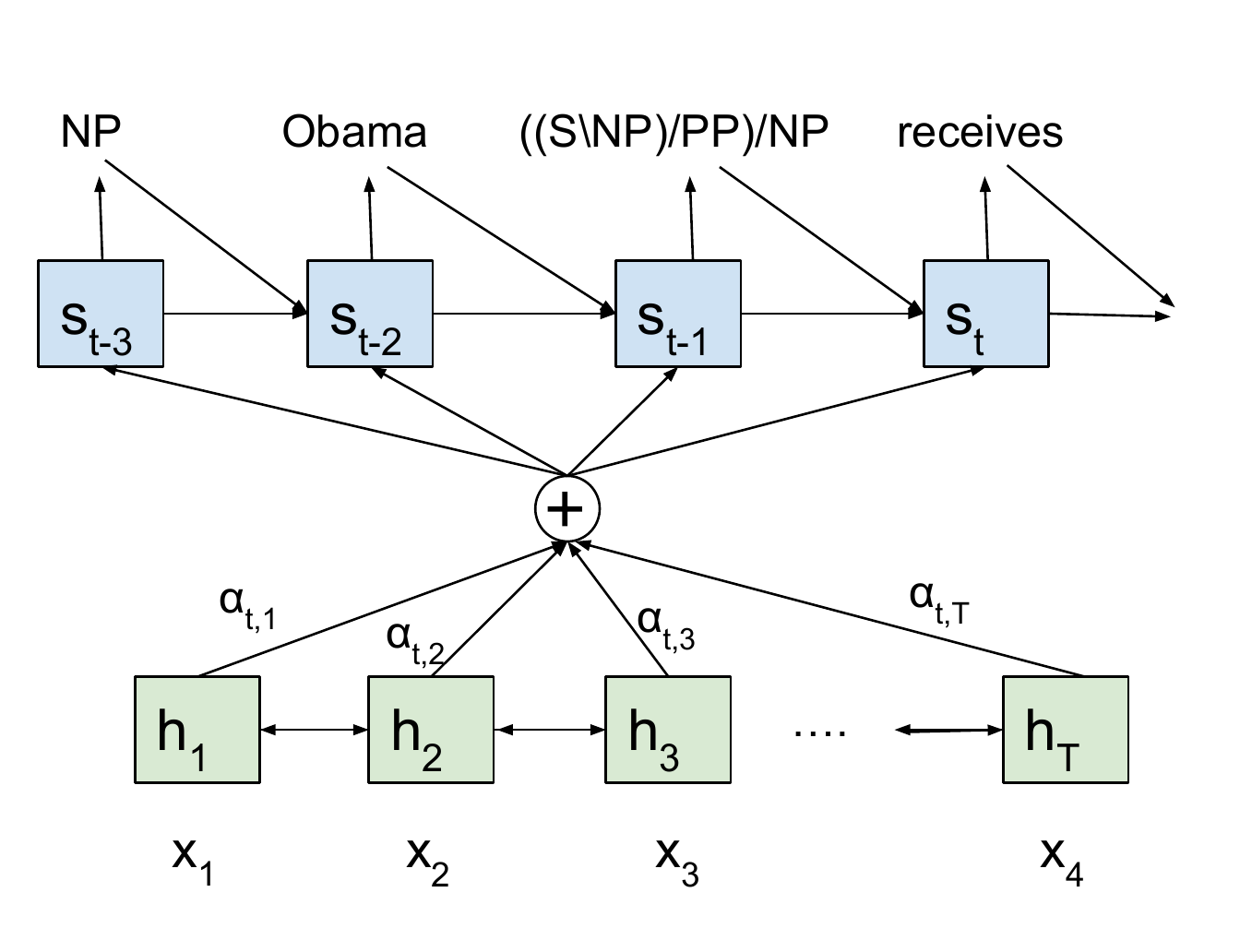} & \includegraphics[width=0.9\columnwidth]{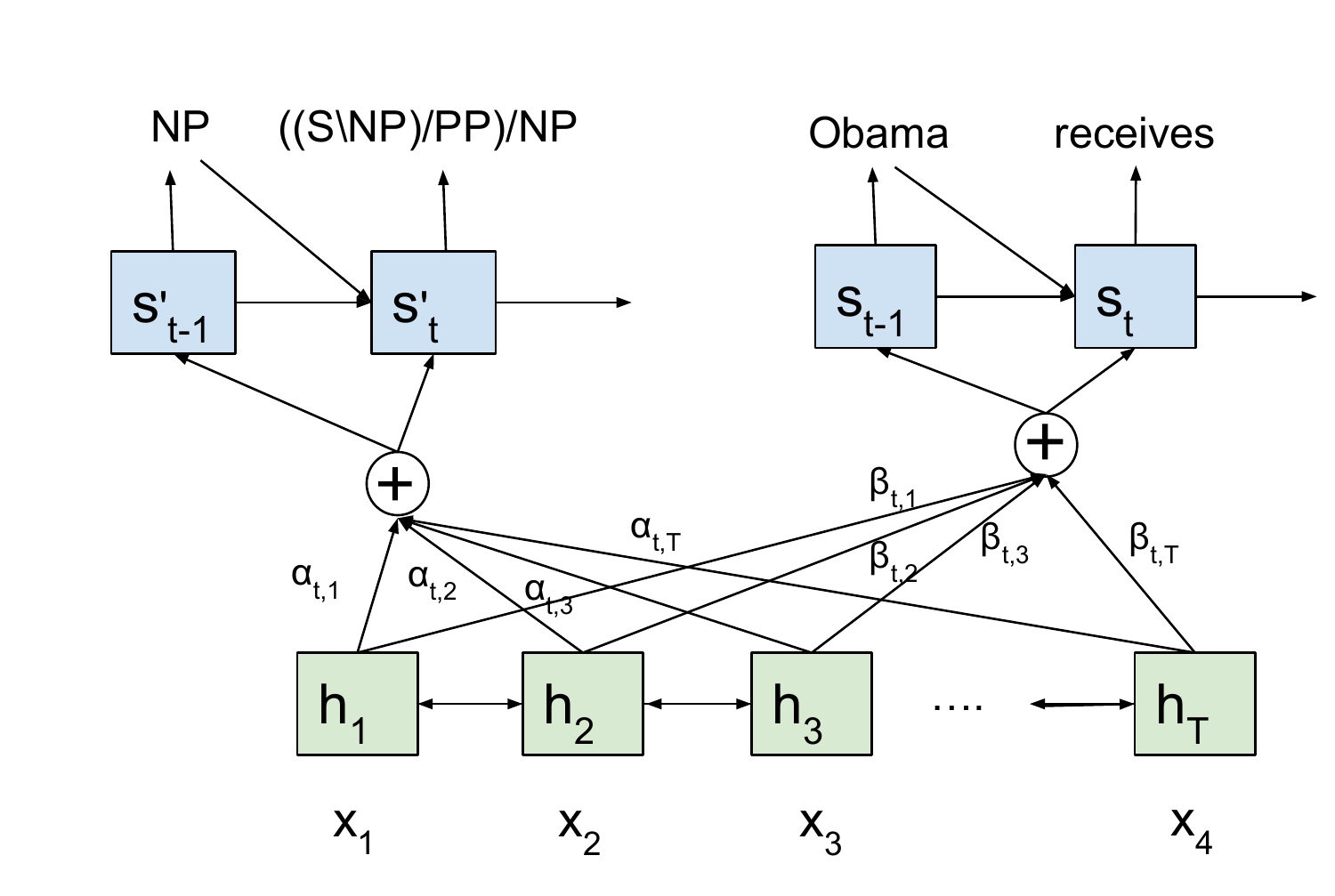} \\
a) & b)

\end{tabular}
\caption{\label{figure:Decoders} Integrating target syntax in the NMT decoder: a) interleaving and b) multitasking. } 
\end{figure*}


\paragraph{Target-side syntax}

When modeling the target-side syntactic information we consider different strategies of coupling the CCG supertags with the translated words in the decoder: interleaving and multitasking with shared encoder. In Figure~\ref{figure:Decoders} we represent graphically the differences between the two strategies and in the next paragraphs we formalize them.

\begin{itemize}[leftmargin=*]
\item \textbf{Interleaving} In this paper we propose a tight integration in the decoder of the syntactic representation and the surface forms. Before each word of the target sequence we include its supertag as an extra token. The new target sequence $y'$ will have the length $2T$, where $T$ is the number of target words.
With this representation, a single decoder learns to predict both the target supertags and the target words conditioned on previous syntactic and lexical context. We do not make changes to the baseline NMT decoder architecture, keeping equations (1) - (6) and the corresponding set of parameters unchanged. Instead, we augment the target vocabulary to include both words and CCG supertags. This results in a shared embedding space and the following probability of the target sequence $y'$, where $y'_j$ can be either a word or a tag: 

\begin{align}
& y' = y^{tag}_{1}, y^{word}_{1}, ....,  y^{tag}_{T}, y^{word}_{T} \\
& p_{y'} = \prod_{j}^{2T} p(y'_{j} | x,y'_{1:j-1})
 \end{align}

 At training time we pre-process the target sequence to add the syntactic annotation and then split only the words into \textit{byte-pair-encoding} (BPE) \citep{sennrich2015bpe} sub-units. At testing time we delete the predicted CCG supertags to obtain the final translation.
   \Cref{figure:SourceCCGRepresentation} gives an example of the target-side representation in the case of interleaving. The supertag \textit{NP} corresponding to the word \textit{Netanyahu} is included only once before the three BPE subunits  \textit{Net+  an+  yahu}.

\item \textbf{Multitasking -- shared encoder} A loose coupling of the syntactic representation and the surface forms can be achieved by co-training a translation model  with a secondary prediction task, in our case CCG supertagging. In the multitask framework \citep{luong2015multi} the encoder part is shared while the decoder is different for each of the prediction tasks: translation and tagging. In contrast to \citeauthor{luong2015multi}, we train a separate attention model for each task and perform multitask learning with target syntax. The two decoders take as input the same source context, represented by the encoder's hidden states $h_{i} = [\overrightarrow{h_{i}};\overleftarrow{h_{i}}]$. However, each task has its own set of parameters associated with the five components of the decoder: $GRU_1$, $ATT$, $cGRU_{att}$, $g$, $softmax$. Furthermore, the two decoders may predict a different number of target symbols, resulting in target sequences of different lengths $T_1$ and $T_2$. This results in two probability distributions over separate target vocabularies for the words and the tags:

\begin{equation}
 p_{y}^{word} = \prod_{j}^{T_1} p(y^{word}_{j} | x,y^{word}_{1:j-1}) 
 \end{equation}
 
 \begin{equation}
 p_{y}^{tag} =  \prod_{k}^{T_2} p(y^{tag}_{k} | x,y^{tag}_{1:k-1})
\end{equation}

The final loss is the sum of the losses for the two decoders: 
\begin{equation}
l = - (log(p^{word}_{y})+ log(p^{tag}_{y}))
\end{equation}
We use EasySRL to label the English side of the parallel corpus with CCG supertags\footnote{We use the same data and annotations for the \textit{interleaving} approach.} instead of using a corpus with gold annotations as in \citet{luong2015multi}. 
\vspace{-0.3em}

\end{itemize}

\paragraph{Source-side syntax -- shared embedding}
While our focus is on target-side syntax, we also experiment with including source-side syntax to show that the two approaches are complementary. 

\citeauthor{sennrich2016linguistic} propose a framework for including source-side syntax as extra features in the NMT encoder.
They extend the model of \citeauthor{bahdanau2015neural} by learning a separate embedding for several source-side features such as the word itself or its part-of-speech. All feature embeddings are concatenated into one embedding vector which is used in all parts of the encoder model instead of the word embedding. When modeling the source-side syntactic information, we include the CCG supertags or dependency labels as extra features.
The baseline features are the subword units obtained using BPE together with the annotation of the subword structure using IOB format by marking if a symbol in the text forms the beginning (B), inside (I), or end (E) of a word. A separate tag (O) is used if a symbol corresponds to the full word. The word level supertag is replicated for each BPE unit. Figure~\ref{figure:SourceCCGRepresentation} gives an example of the source-side feature representation.

\section{Experimental Setup and Evaluation}
\subsection{Data and methods}

We train the neural MT systems on all the parallel data available at WMT16 \citep{bojar-EtAl:2016:WMT16} for the German$\leftrightarrow$English and Romanian$\leftrightarrow$English language pairs. 
The English side of the training data is annotated with CCG lexical tags\footnote{The CCG tags include features such as the verb tense (e.g. [ng] for continuous form) or the sentence type (e.g. [pss] for passive).} using EasySRL \citep{lewis:2015} and the available pre-trained model\footnote{\url{https://github.com/uwnlp/EasySRL}}. 
Some longer sentences cannot be processed by the parser and therefore we eliminate them from our training and test data. We report the sentence counts for the filtered data sets in Table~\ref{table:data}.  Dependency labels are annotated with ParZU \citep{sennrich13c} for German and SyntaxNet \citep{andor-EtAl:2016:P16-1} for Romanian.

\begin{table}[t]
\begin{center}
\begin{adjustbox}{max width=\columnwidth}
\begin{tabular}{l | r | r | r }

  & train & dev & test \\  \hline
  DE-EN & 4,468,314 &  2,986 & 2,994  \\ 
  RO-EN & 605,885 & 1,984 & 1,984  \\ 

\end{tabular}
\end{adjustbox}
\end{center}
\caption{\label{table:data} Number of sentences in the training, development and test sets.}
\vspace{-1em}
\end{table}

All the neural MT systems are attentional encoder-decoder networks \citep{bahdanau2015neural} as implemented in the Nematus toolkit \citep{sennrich-EtAl:2017}.\footnote{https://github.com/rsennrich/nematus} We use similar hyper-parameters to those reported by \cite{sennrich-haddow-birch:2016:WMT, sennrich2016linguistic} with minor modifications: we used mini-batches of size 60 and Adam optimizer \citep{kingma2014adam}. We select the best single models according to \bleu on the development set and use the four best single models for the ensembles. 

To show that we report results over strong baselines, table~\ref{table:baselines} compares the scores obtained by our baseline system to the ones reported in \citet{sennrich-haddow-birch:2016:WMT}. We normalize diacritics\footnote{There are different encodings for letters with cedilla (\c s,\c t) used interchangeably throughout the corpus.  \url{https://en.wikipedia.org/wiki/Romanian_alphabet\#ISO_8859} } for the English$\rightarrow$Romanian test set. We did not remove or normalize Romanian diacritics for the other experiments reported in this paper. Our baseline systems are generally stronger than \citet{sennrich-haddow-birch:2016:WMT} due to training with a different optimizer for more iterations.

\begin{table}[ht]
\begin{center}
\begin{adjustbox}{max width=1\textwidth}
\begin{tabular}{ l |  c | c } 

  & This work & Sennrich et. al \\ \hline
 DE$\rightarrow$EN & 31.0 & 28.5   \\ 
 EN$\rightarrow$DE  & 27.8  & 26.8 \\
 RO$\rightarrow$EN  & 28.0 & 27.8   \\   
 EN$\rightarrow$RO$^1$  & 25.6  & 23.9   \\  
 
\end{tabular}
\end{adjustbox}
\end{center}
\caption{\label{table:baselines} Comparison of baseline systems in this work and in  \citet{sennrich-haddow-birch:2016:WMT}. Case-sensitive \bleu scores reported over newstest2016 with  \textit{mteval-13a.perl}. $^1$Normalized diacritics.}
\vspace{-1em}
\end{table}

During training we validate our models with \bleu \citep{BLEU} on development sets: newstest2013 for German$\leftrightarrow$English and newsdev2016 for Romanian$\leftrightarrow$English. We evaluate the systems on newstest2016 test sets for both language pairs and use bootstrap resampling \citep{riezler-maxwell:2005:MTSumm} to test statistical significance. 
We compute \bleu with \textit{multi-bleu.perl} over tokenized sentences both on the development sets, for early stopping, and on the test sets for evaluating our systems.

Words are segmented into sub-units that are learned jointly for source and target using BPE \citep{sennrich2015bpe}, resulting in a vocabulary size of 85,000. The vocabulary size for CCG supertags was 500. 


\begin{table*}[ht]
\begin{center}
\begin{adjustbox}{max width=1\textwidth}
\begin{tabular}{ l |  l | l |   c | c    ||  c | c} 

 & & & \multicolumn{2}{c ||}{German$\rightarrow$English} &  \multicolumn{2}{c }{Romanian$\rightarrow$English} \\ \hline
 model & syntax & strategy   & single & ensemble & single & ensemble\\ \hline 
 NMT & - & -    & 31.0 &  32.1 \ \  \       & 28.1 & 28.4 \ \ \  \\   
 SNMT & target -- CCG & interleaving     & 32.0  & \textbf{32.7}*\ \  &  29.2 & \textbf{29.3}**  \\  
 Multitasking & target -- CCG & shared encoder & 31.4 & 32.0  \ \  \  & 28.4 & 29.0* \ \   \\ \hline  \hline  
 SNMT & source -- dep & shared embedding  &  31.4  & 32.2 \ \  \ & 28.2 & 28.9 \ \ \   \\  
   & \ + target -- CCG & \ + interleaving  &  32.1 & \textbf{33.0}**  & 29.1 & \textbf{29.6}**   \\ 

\end{tabular}
\end{adjustbox}
\end{center}
\caption{\label{table:TargetSyntax}  Experiments with target-side syntax  for German$\rightarrow$English and  Romanian$\rightarrow$English. \bleu scores reported for baseline NMT, syntax-aware NMT (SNMT) and multitasking. The SNMT system is also combined with source dependencies. Statistical significance is indicated with * $p<0.05$ and ** $p<0.01$, when comparing against the NMT baseline.}  
\end{table*}


For the experiments with source-side features we use the BPE sub-units and the IOB tags as baseline features. We keep the total word embedding size fixed to 500 dimensions. We allocate 10 dimensions for dependency labels when using these as source-side features and when using source-side CCG supertags we allocate 135 dimensions.

The \textit{interleaving} approach to integrating target syntax increases the length of the target sequence. Therefore,  at training time, when adding the CCG supertags in the target sequence we increase the maximum length of sentences from 50 to 100. On average, the length of English sentences for newstest2013 in BPE representation is 22.7, while the average length when adding the CCG supertags is 44. Increasing the length of the target recurrence results in larger memory consumption and slower training.\footnote{Roughly 10h30 per 100,000 sentences (20,000 batches) for SNMT compared to 6h for NMT.}. At test time, we obtain the final translation by post-processing the predicted target sequence to remove the CCG supertags.

\subsection{Results}

In this section, we first evaluate the syntax-aware NMT model (SNMT) with target-side CCG supertags as compared to the baseline NMT model described in the previous section \citep{bahdanau2015neural, sennrich-haddow-birch:2016:WMT}.
We show that our proposed method for tightly coupling target syntax via \textit{interleaving}, improves translation for both German$\rightarrow$English and Romanian$\rightarrow$English while the \textit{multitasking} framework does not. Next, we show that SNMT with target-side CCG supertags can be complemented with source-side dependencies, and that combining both types of syntax brings the most improvement. 
Finally, our experiments with source-side CCG supertags confirm that global syntax can improve translation either as extra information in the encoder or in the decoder.

\paragraph{Target-side syntax}

We first evaluate the impact of target-side CCG supertags on overall translation quality. In Table~\ref{table:TargetSyntax} we report results for German$\rightarrow$English, a high-resource language pair, and for Romanian$\rightarrow$English, a low-resource language pair. 
We report \bleu scores for both the best single models and ensemble models. However, we will only refer to the results with ensemble models since these are generally better.

The SNMT system with target-side syntax improves \bleu scores by 0.9 for Romanian$\rightarrow$English and by 0.6 for  German$\rightarrow$English. Although the training data for German$\rightarrow$English is large, the CCG supertags still improve translation quality.
These results suggest that the baseline NMT decoder benefits from modeling the global syntactic information locally via supertags. 

Next, we evaluate whether there is a benefit to tight coupling between the target word sequence and syntax, as apposed to loose coupling. We compare our method of \textsl{interleaving} the CCG supertags with \textsl{multitasking}, which predicts target CCG supertags as a secondary task.
The results in Table~\ref{table:TargetSyntax} show that the multitask approach does not improve \bleu scores for German$\rightarrow$English, which exhibits long distance word reordering. For Romanian$\rightarrow$English, which exhibits more local word reordering, multitasking improves \bleu by 0.6 relative to the baseline. 
In contrast, the \textit{interleaving} approach improves translation quality for both language pairs and to a larger extent. Therefore, we conclude that a tight integration of the target syntax and word sequence is important. Conditioning the prediction of words on their corresponding CCG supertags is what sets SNMT apart from the multitasking approach.


\paragraph{Source-side and target-side syntax} We now show that our method for integrating target-side syntax can be combined with the framework of \citet{sennrich2016linguistic} for integrating source-side linguistic information, leading to further improvement in translation quality. We evaluate the syntax-aware NMT system, with CCG supertags as target-syntax and dependency labels as source-syntax. While the dependency labels do not encode global syntactic information, they disambiguate the grammatical function of words. Initially, we had intended to use global syntax on the source-side as well for German$\rightarrow$English, however the German CCG tree-bank is still under development. 

From the results in Table~\ref{table:TargetSyntax} we first observe that for German$\rightarrow$English the source-side dependency labels improve \bleu by only 0.1, while Romanian$\rightarrow$English sees an improvement of 0.5. Source-syntax may help more for Romanian$\rightarrow$English because the training data is smaller and the word order is more similar between the source and target languages than it is for German$\rightarrow$English. 

For both language pairs, target-syntax improves translation quality more than source-syntax. However, target-syntax is complemented by source-syntax when used together, leading to a final improvement of 0.9 \bleu points for German$\rightarrow$English  and 1.2 \bleu points for  Romanian$\rightarrow$English.

\begin{figure*}[ht]
\begin{tabular}{c c}
 \includegraphics[width=1\columnwidth]{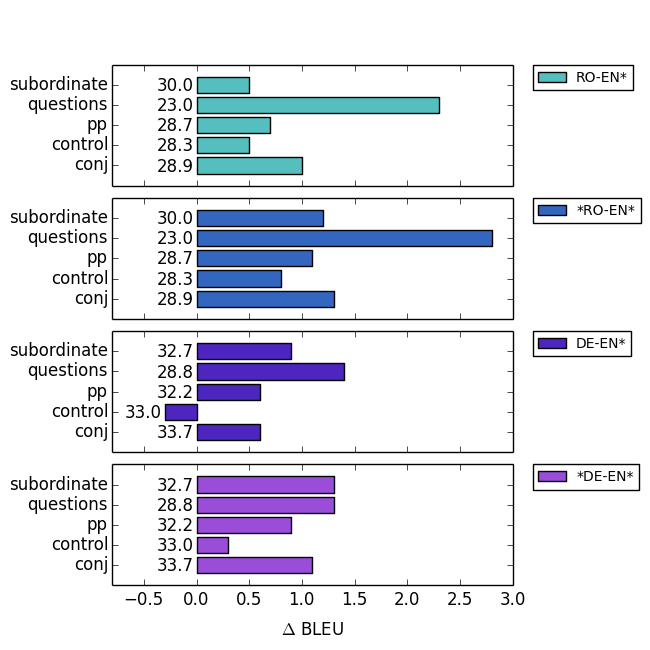} & \includegraphics[width=1\columnwidth]{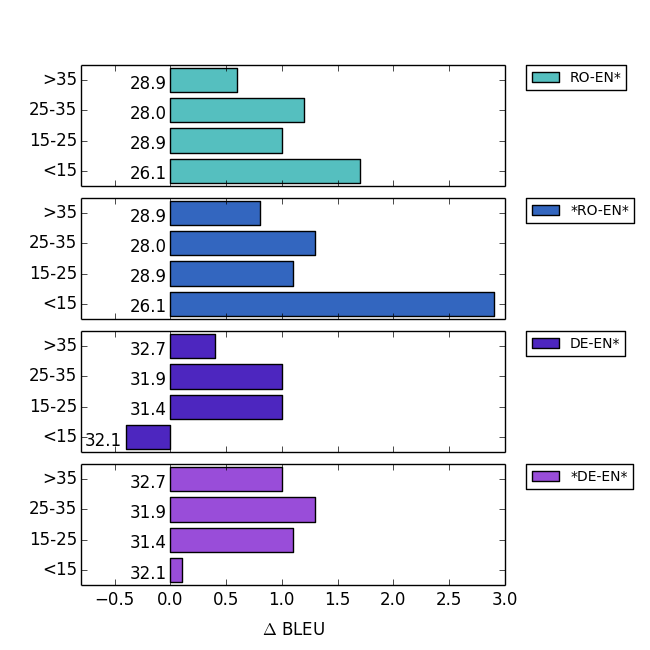} \\
a) & b)

\end{tabular}
\caption{\label{figure:TargetLengthLinguisticBleu} Difference in \bleu points between SNMT and NMT,  relative to baseline NMT scores, with respect to a) linguistic constructs and b) sentence lengths. The numbers attached to the bars represent the \bleu score for the baseline NMT system. The symbol * indicates that syntactic information is used on the target (eg. de-en*),  or both on the source and target (eg. *de-en*) } 
\end{figure*}

Finally, we show that CCG supertags are also an effective representation of global-syntax when used in the encoder. 
In Table~\ref{table:SourceSyntax}  we present results for using CCG supertags as source-syntax in the embedding layer. Because we have CCG annotations only for English, we reverse the translation directions and report \bleu scores for English$\rightarrow$German and English$\rightarrow$Romanian. The \bleu scores reported are for the ensemble models over newstest2016.  

\begin{table}[h]
\begin{center}
\begin{adjustbox}{max width=1\textwidth}
\begin{tabular}{ l | l | c  | c}

 model & syntax  & EN$\rightarrow$DE & EN$\rightarrow$RO \\ \hline 
 NMT & -   & 28.3 \ \  &  25.6 \ \   \\  
 SNMT & source -- CCG   & \textbf{29.0}*     & \textbf{26.1}* \\

\end{tabular}
\end{adjustbox}
\end{center}
\caption{\label{table:SourceSyntax} Results for English$\rightarrow$German and English$\rightarrow$Romanian with source-side syntax. The SNMT system uses the CCG supertags of the source words in the embedding layer. *$p<0.05$.}  
\vspace{-1em}
\end{table}

For English$\rightarrow$German \bleu increases by 0.7 points and for English$\rightarrow$Romanian by 0.5 points. In contrast, \citet{sennrich2016linguistic} obtain an improvement of only 0.2 for English$\rightarrow$German using dependency labels which encode only the grammatical function of words. These results confirm that representing global syntax in the encoder provides complementary information that the baseline NMT model is not able to learn from the source word sequence alone.

\subsection{Analyses by sentence type}

In this section, we make a finer grained analysis of the impact of target-side syntax by looking at a breakdown of \bleu scores with respect to different linguistic constructions and sentence lengths\footnote{Document-level \bleu is computed over each subset of sentences.}. 

We classify sentences into different linguistic constructions based on the CCG supertags that appear in them, e.g., the presence of category \cf{(NP\backslash NP)/(S/NP)} indicates a subordinate construction. 
Figure~\ref{figure:TargetLengthLinguisticBleu} a) shows the difference in \bleu points between the syntax-aware NMT system  and the baseline NMT system for the following linguistic constructions: coordination \textit{(conj)}, control and raising \textit{(control)}, prepositional phrase attachment \textit{(pp)}, questions and subordinate clauses \textit{(subordinate)}. In the figure we use the symbol ``*'' to indicate that syntactic information is used on the target (eg. de-en*),  or both on the source and target (eg. *de-en*).
We report the number of sentences for each category in Table~\ref{table:CCGSentenceCount}.

\begin{table}[h]
\begin{center}
\begin{adjustbox}{max width=1\textwidth}
\begin{tabular}{ l |  c | c | c | c | c}

& sub. & qu. & pp & contr. & conj  \\ \hline
RO$\leftrightarrow$EN & 742 & 90 & 1,572 & 415 & 845 \\
DE$\leftrightarrow$EN & 936 & 114 & 2,321 & 546 & 1,129 \\

\end{tabular}
\end{adjustbox}
\end{center}
\caption{\label{table:CCGSentenceCount} Sentence counts for different linguistic constructions.} 
\end{table}

With target-syntax, we see consistent improvements across all linguistic constructions for Romanian$\rightarrow$English and across all but \textit{control and raising} for German$\rightarrow$English.
In particular, the increase in \bleu scores for the \textit{prepositional phrase} and  \textit{subordinate} constructions suggests that target word order is improved. 

For German$\rightarrow$English, there is a small decrease in \bleu for the \textit{control and raising} constructions when using target-syntax alone. However, source-syntax adds complementary information to target-syntax, resulting in a small improvement for this category as well.
Moreover, combining source and target-syntax increases translation quality across all linguistic constructions as compared to NMT and SNMT with target-syntax alone. For Romanian$\rightarrow$English, combining source and target-syntax brings an additional improvement of 0.7 for \textit{subordinate} constructs and 0.4 for \textit{prepositional phrase attachment}. For German$\rightarrow$English, on the same categories, there is an additional improvement of 0.4 and 0.3 respectively. 
 Overall, \bleu scores improve by more than 1 \bleu point for most linguistic constructs and for both language pairs.

Next, we compare the systems with respect to sentence length. Figure~\ref{figure:TargetLengthLinguisticBleu} b) shows the difference in \bleu points between the syntax-aware NMT system and the baseline NMT system with respect to the length of the source sentence measured in BPE sub-units. We report the number of sentences for each category in Table~\ref{table:LenSentenceCount}. 

\begin{figure*}[ht]
\begin{center}
\begin{adjustbox}{max width=1\textwidth}
\begin{tabular}{ l  l}

& \textbf{DE - EN Question} \\ \hline
 Source & Oder wollen Sie herausfinden , \textbf{\"uber} was andere reden ?    \\
 Ref. & Or do you want to find out what others are talking \textbf{about} ?  \\
NMT  & Or would you like to find out \textbf{about} what others are talking \textbf{about} ? \\
 SNMT &  Or do you want to find out what$_{NP/(S[dcl]/NP)}$ others are$_{(S[dcl]\backslash NP)/(S[ng]\backslash NP)}$ talking$_{(S[ng]\backslash NP)/PP}$ \textbf{about}$_{PP/NP}$ ?
 
  \\
 & \\
 & \textbf{DE - EN Subordinate} \\ \hline
 Source & ...dass die Polizei jetzt sagt , ..., und dass Lamb in seinem Notruf \textbf{Prentiss zwar als seine Frau bezeichnete} ...\\
 Ref. & ...that police are now saying ..., and that while Lamb \textbf{referred to Prentiss as his wife} in the 911 call ...   \\
NMT  & ...police are now saying  ..., and that in his emergency call \textbf{Prentiss he called his wife} ...\\
 SNMT & ...police are now saying ..., and that lamb , in his emergency call , \textbf{described$_{((S[dcl]\backslash NP)/PP)/NP}$ Prentiss as his wife} ....

\end{tabular}
\end{adjustbox}
\end{center}
\caption{\label{figure:TargetFactorsExamplesShort} Comparison of baseline NMT and SNMT with target syntax for German$\rightarrow$English.} 

\end{figure*}

\begin{table}[h]
\begin{center}
\begin{adjustbox}{max width=1\textwidth}
\begin{tabular}{ l |  c | c | c | c }

& $<$15 & 15-25 & 25-35 & $>$35  \\ \hline
RO$\leftrightarrow$EN & 491 & 540 & 433 & 520 \\
DE$\leftrightarrow$EN & 918 & 934 & 582 & 560\\

\end{tabular}
\end{adjustbox}
\end{center}
\caption{\label{table:LenSentenceCount} Sentence counts for different sentence lengths.} 
\end{table}

With target-syntax, we see consistent improvements across all sentence lengths for Romanian$\rightarrow$English and across all but short sentences for German$\rightarrow$English.
For German$\rightarrow$English there is a decrease in \bleu for sentences up to 15 words. Since the German$\rightarrow$English training data is large, the baseline NMT system learns a good model for short sentences with local dependencies and without subordinate or coordinate clauses. Including extra CCG supertags increases the target sequence without adding information about complex linguistic phenomena. However, when using both source and target syntax, the effect on short sentences disappears. For Romanian$\rightarrow$English there is also a large improvement on short sentences when combining  source and target syntax: 2.9 \bleu points compared to the NMT baseline and 1.2 \bleu points compared to SNMT with target-syntax alone.

With both source and target-syntax, translation quality increases across all sentence lengths as compared to NMT and SNMT with target-syntax alone.
For German$\rightarrow$English sentences that are more than 35 words, we see again the effect of increasing the target sequence by adding CCG supertags. Target-syntax helps, however \bleu improves by only 0.4, compared to 0.9 for sentences between 15 and  35 words. With both source and target syntax, \bleu improves by 0.8 for sentences with more than 35 words. For Romanian$\rightarrow$English we see a similar result for sentences with more than 35 words: target-syntax improves \bleu by 0.6, while combining source and target syntax improves \bleu by 0.8. These results confirm as well that source-syntax adds complementary information to target-syntax and mitigates the problem of increasing the target sequence. 

\subsection{Discussion}

Our experiments demonstrate that target-syntax improves translation for two translation directions: German$\rightarrow$English and Romanian$\rightarrow$English. Our proposed method predicts the target words together with their CCG supertags.

Although the focus of this paper is not improving CCG tagging, we can also measure that SNMT is accurate at predicting CCG supertags. We compare the CCG sequence predicted by the SNMT models with that predicted by EasySRL and obtain the following accuracies: 93.2 for Romanian$\rightarrow$English, 95.6 for  German$\rightarrow$English, 95.8 for  German$\rightarrow$English with both source and target syntax.\footnote{The multitasking model predicts a different number of CCG supertags than the number of target words. For the sentences where these numbers match, the CCG supetagging accuracy is 73.2.} 

We conclude by giving a couple of examples in Figure~\ref{figure:TargetFactorsExamplesShort} for which the SNMT system with target syntax produced more grammatical translations than the baseline NMT system.

In the example \textbf{DE-EN Question}  the baseline NMT system translates the preposition \textit{``\"uber''} twice as \textit{``about''}. The SNMT system with target syntax predicts the correct CCG supertag for \textit{``what''} which expects to be followed by a sentence and not a preposition: \cf{NP/(S[dcl]/NP)}. Therefore the SNMT correctly re-orders the preposition \textit{``about''} at the end of the question.

In the example \textbf{DE-EN Subordinate} the baseline NMT system fails to correctly attach  \textit{``Prentiss''} as an object and \textit{``his wife''} as a modifier to the verb \textit{``called (bezeichnete)''} in the subordinate clause. In contrast the SNMT system predicts the correct sub-categorization frame of the verb \textit{``described''} and correctly translates the entire predicate-argument structure.
  


\section{Conclusions}
\label{conclusions}
This work introduces a method for modeling explicit target-syntax in a neural machine translation system, by interleaving target words with their corresponding CCG supertags. Earlier work on syntax-aware NMT mainly modeled syntax in the encoder, while our experiments suggest modeling syntax in the decoder is also useful. 
Our results show that a tight integration of syntax in the decoder improves translation quality for both German$\rightarrow$English and Romanian$\rightarrow$English language pairs, more so than a loose coupling of target words and syntax as in multitask learning. 
Finally, by combining our method for integrating target-syntax with the framework of \citet{sennrich2016linguistic} for source-syntax we obtain the most improvement over the baseline NMT system: 0.9 \bleu for German$\rightarrow$English and 1.2 \bleu for Romanian$\rightarrow$English. In particular, we see large improvements for longer sentences involving syntactic phenomena such as subordinate and coordinate clauses and prepositional phrase attachment. In future work, we plan to evaluate the impact of target-syntax when translating into a morphologically rich language, for example by using the Hindi CCGBank~\citep{AmbatiCCGBank2016}.

\section*{Acknowledgements}

We thank the anonymous reviewers for their comments and suggestions. This project has received funding from
the European Union's Horizon 2020 research and innovation programme under grant agreements 644402 (HimL), 644333 (SUMMA) and 645452 (QT21).



\addcontentsline{toc}{section}{References}
\bibliographystyle{emnlp_natbib}
\bibliography{acl2015}

\end{document}